\def\year{2021}\relax
\documentclass[letterpaper]{article} % DO NOT CHANGE THIS
\usepackage{aaai21}  % DO NOT CHANGE THIS
\usepackage{times}  % DO NOT CHANGE THIS
\usepackage{helvet} % DO NOT CHANGE THIS
\usepackage{courier}  % DO NOT CHANGE THIS
\usepackage[hyphens]{url}  % DO NOT CHANGE THIS
\usepackage{graphicx} % DO NOT CHANGE THIS
\usepackage{tikz}
\usepackage{natbib}  % DO NOT CHANGE THIS AND DO NOT ADD ANY OPTIONS TO IT
\usepackage{caption} % DO NOT CHANGE THIS AND DO NOT ADD ANY OPTIONS TO IT
\usepackage{xfrac}
\usepackage{adjustbox}
\usepackage{multirow}
\usepackage{multicol}
\usepackage[linesnumbered, ruled,vlined]{algorithm2e}
\usepackage{booktabs}

\title{The Role of Isomorphism Classes in Multi-Relational Datasets}
\author{
    Vijja Wichitwechkarn*\textsuperscript{\rm 1}, Ben Day*\textsuperscript{\rm 2}, Cristian Bodnar*\textsuperscript{\rm 2}, Matthew Wales\textsuperscript{\rm 3}, Pietro Li\`{o}\textsuperscript{\rm 2}\\
}
\affiliations{
    \textsuperscript{\rm 1}Department of Physics, University of Cambridge\\
    
    \textsuperscript{\rm 2}Computer Laboratory, University of Cambridge\\
    
    \textsuperscript{\rm 3}Department of Mathematics, University of Cambridge\\
    
    \{bjd39,cb2015\}@cam.ac.uk
    
}

\begin{document}

\maketitle

\begin{abstract}
Multi-interaction systems abound in nature, from colloidal suspensions to gene regulatory circuits. These systems can produce complex dynamics and graph neural networks have been proposed as a method to extract underlying interactions and predict how systems will evolve. The current training and evaluation procedures for these models through the use of synthetic multi-relational datasets however are agnostic to interaction network isomorphism classes, which produce identical dynamics up to initial conditions. We extensively analyse how isomorphism class awareness affects these models, focusing on neural relational inference (NRI) models, which are unique in explicitly inferring interactions to predict dynamics in the unsupervised setting. Specifically, we demonstrate that isomorphism leakage overestimates performance in multi-relational inference and that sampling biases present in the multi-interaction network generation process can impair generalisation. To remedy this, we propose isomorphism-aware synthetic benchmarks for model evaluation. We use these benchmarks to test generalisation abilities and demonstrate the existence of a threshold sampling frequency of isomorphism classes for successful learning. In addition, we demonstrate that isomorphism classes can be utilised through a simple prioritisation scheme to improve model performance, stability during training and reduce training time.
\end{abstract}

\section{Introduction}
We focus on the task of predicting the dynamics of simple many-body multi-interaction systems, a first step towards scaling to more complex dynamical systems. A variety of approaches have been developed to tackle variants of this problem, including predicting the trajectories of particles given the underlying interaction network \cite{Battaglia:example}, learning to simulate complex physics with graph networks \cite{Sanchez-Gonzalez:example}, and applying constraints from Lagrangian dynamics to learn a physics model \cite{Lutter:Example}. We will focus on approaches that both predict trajectories and infer relations in the system, neural relational inference (NRI) \cite{Kipf:NRI} and factorised neural relation inference (fNRI) \cite{Webb:fNRI}. These are unsupervised models which explicitly infer the underlying interactions of a system to predict the resulting dynamics. This structure is akin to an interpretable theory and the predictions it makes, with the aim of being a more valuable research tool compared to inscrutable blackboxes. The investigations conducted will also be relevant to synthetic multi-relational datasets from other settings \cite{Sinha:logical_generalisation}.

\begin{figure*}[h!]
    \centering
    \includegraphics[trim = 6 65 22 6, clip, width=\textwidth]{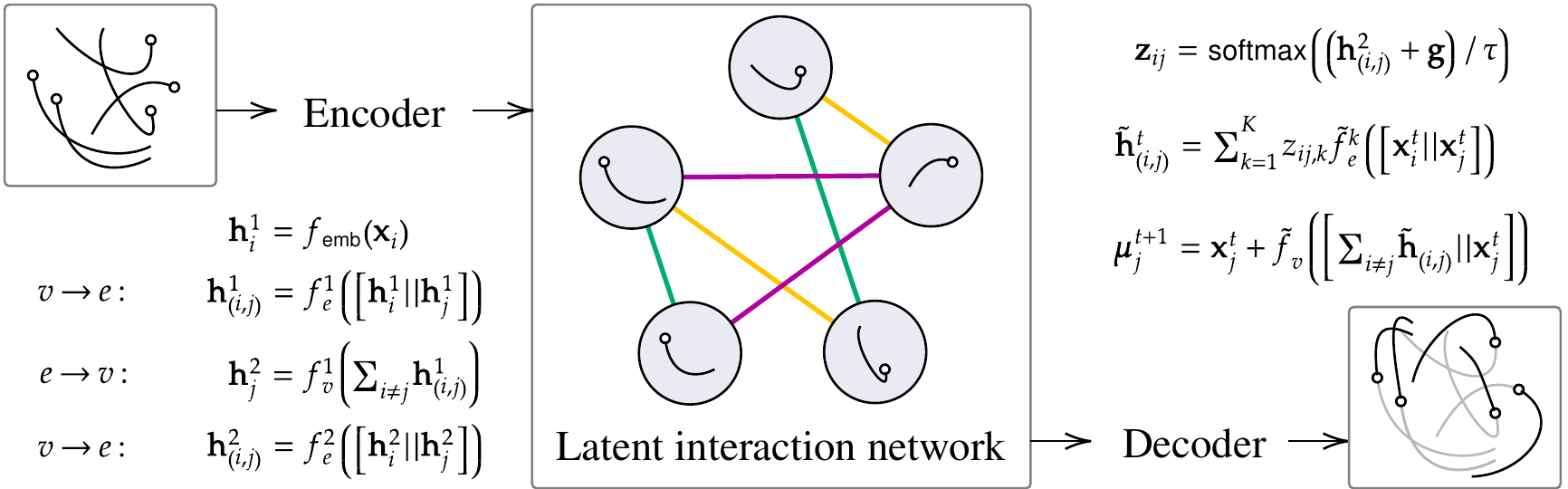}
    
    \caption{Model overview. The encoder embeds trajectories ($\mathbf{x}$) and, using vertex-to-edge ($v \rightarrow e$) and edge-to-vertex ($e \rightarrow v$) message-passing operations, produces the latent interaction network. The sampled edges ($\mathbf{z}$) modulate pairwise functions ($\tilde{f}^k_e$) in the decoder that can be associated with forces in classical physics. A function of the net resultant `force' (sum over $k$) is used to update the mean position using a skip-connection. $[x||y]$ indicates concatenation.}
    \label{fig:model_overview}
    
\end{figure*}

Despite designing the model architecture around the potential value of explicitly inferring interactions, little attention is paid to the structure of multiplex interaction networks or their sampling distribution in training and evaluation routines. There are many non-obvious results in the field of random graph theory, perhaps the most well-known being the percolation transition where, above a threshold connectivity, it is expected that a single component will come to encompass the entire graph \cite{Erdos:1, Erdos:2}. Effects such as these can bias the generation of the synthetic data used to train these models, hampering generalisation and causing performance to be overestimated.

A second missing component is the scientific process by which experiments are formulated to test the edge of current theories: areas well within the understood domain provide little new information, whereas regions far beyond our understanding are often too poorly explained to allow insight to be gained from results. Scientific progress is driven by this almost antagonistic relationship between theorists and experimentalists. This is absent from the training procedure of these models, where examples are treated without consideration as to the model's current performance.

We incorporate these concerns into better synthetic multi-relation dataset generation and a new training procedure in this work. To do this, we first analyse the structure of interaction networks (Section \ref{section:interaction}), exposing non-intuitive results in the distribution of multiplex isomorphism classes and exploring how generation methods can incur a bias and leak over generated datasets. We demonstrate how these biases impact training and the overestimation of model performance arising due to leakage. We also (tentatively) present a novel fast algorithm for the construction of the set of non-isomorphic interaction networks, with a proof of correctness\footnote{To the best of our knowledge, following a thorough literature review and consultation with domain experts, the algorithm and proof are original work, though we welcome any suggestions of prior-art.}. We then present isomorphism-aware benchmarks which were used to evaluate the model's performance and identify the presence of a threshold sampling frequency of isomorphism classes for successful learning (Section \ref{section:model}). Finally, we show that incorporating isomorphism awareness via priority sampling improves model performance, stability during training, and significantly reduces training time (Section \ref{subsec:prioritised}).

\section{Model background}
We present a brief overview of the task formulation and state-of-the-art approaches, the Neural Relational Inference (NRI) model \cite{Kipf:NRI} and its derivative, the Factorised Neural Relational Inference (fNRI) model \cite{Webb:fNRI}. We do not make any architectural modifications to the original models.

\paragraph{Problem statement} The primary task is the reconstruction (or evolution) of trajectories of particles in an interacting system, represented as a sequence of feature vectors over particles, $\mathbf{x}_i=\{\mathbf{x}_i^0,...,\mathbf{x}_i^T\}$, and time, $\mathbf{x}^t=\{\mathbf{x}_1^t,...,\mathbf{x}_N^t\}$, with neither access to nor supervision from the ground truth interaction network.

\paragraph{Model formulation} Both the NRI and fNRI are formulated as variational-autoencoders (VAEs) with observed trajectories being encoded as a latent interaction network that determines the output trajectory evolution for some initial conditions. Architecturally, the models are graph neural networks that use message passing in the encoder and decoder. The encoder embeds each particle's trajectory then, through a series of vertex-to-edge and edge-to-vertex message passing operations, produces an edge-embedding between each pair of particles. The models differ in the dimensionality and meaning of the edge-embedding: the NRI uses a one-hot $2^k$-dimensional vector with a separate edge-type for each interaction and combination of interactions; the fNRI uses a multi-categorical vector of length $2k$ where different edge types exist only for different interactions. The decoder samples the latent interaction network to modulate the message-passing between particles, corresponding to the transmission of force-carrying particles. Figure \ref{fig:model_overview} presents the model diagrammatically.

Though the fNRI outperforms the original NRI, that the models differ in their handling of the latent interaction networks makes them both relevant to our analysis of the impact that interaction network sampling has on performance estimation.

\section{Isomorphism analysis}\label{section:interaction}
\begin{figure}[h]
\begin{center}
\includegraphics[trim = 8 8 175 8, clip, width=\columnwidth]{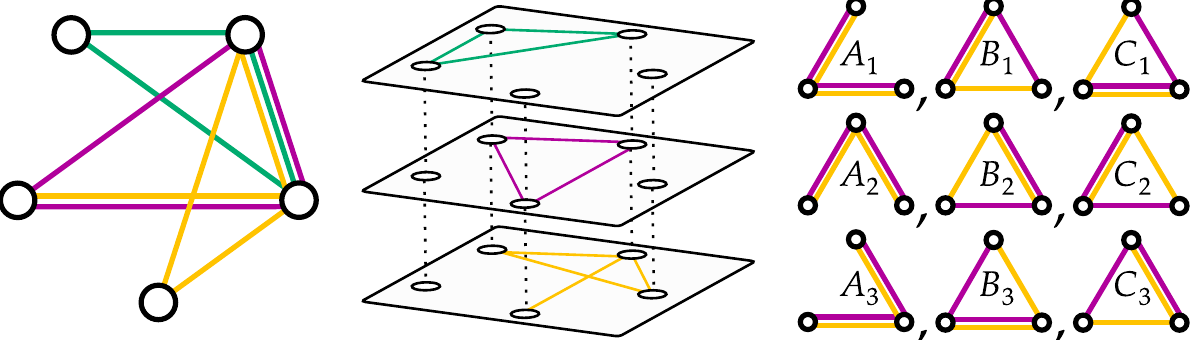}
\end{center}
\caption{Edge-coloured multi-graph (left) and multiplex network (centre) representations of an interacting system; and the combinations of a pair of basis graphs (right). $\{A_1,A_2,A_3\}$ form an equivalence class based on rotations, similarly for $\{B_1,B_2,B_3\}$ and $\{C_1,C_2,C_3\}$, respectively. In addition, the $B$ and $C$ graphs collectively form an equivalence class (analogous to reflections). If the basis graphs were combined at random, $\{B \equiv C\}$ would be selected twice as frequently as $A$.}
\label{figure:purple_graphs}
\end{figure}

In this section we analyse interaction networks through their isomorphism classes, investigate the sampling biases that arise from standard multi-interaction network generation processes, and show how information can leak between datasets through isomorphisms. The influence of the bias and leakage on performance evaluation is presented. Here we focus our analysis on five particles interacting via ideal-springs, finite-springs, and a charge force in two dimensions, as in the original work \cite{Webb:fNRI}.

\subsection{Isomorphism classes}
The set of possible interaction networks for some combination of interactions can be partitioned into isomorphism classes which, up to initial conditions, result in identical particle dynamics. In this sense the isomorphism classes form the set of `unique networks' that can be generated.

\begin{figure}[ht]
\begin{center}
\includegraphics[trim = 0 175 25 0, clip, width=\columnwidth]{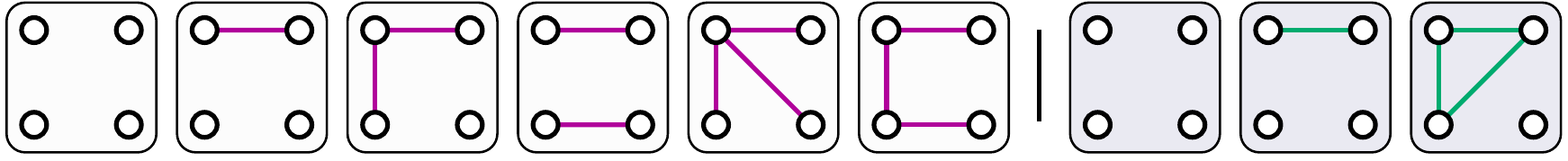}
\end{center}
\caption{Sparse complement basis networks for pairwise (left, purple) and collective (right, green) interactions for 4 particles. There are 11 graphs with 4 vertices, but for pairwise we can use the 6 shown and for collective just 3 will suffice, without loss of generality. This significantly reduces the computational cost of finding the set of unique multiplex networks. Note that collective interactions always consist of a single fully-connected component with the other particles being isolated (particles that are charged interact with all other charged particles).}
\label{figure:basis_graphs}
\end{figure}

\paragraph{Basis networks} We can simplify our analysis by first considering the isomorphism classes of the base interactions separately, as the multiplex network\footnote{A multiplex network is a vertex-aligned multilayer graph. A vertex exists in every layer and is only connected to itself across layers.} is itself formed of the combinations and node-permutations of these (see Figure \ref{figure:purple_graphs}). In our experiments the interactions can either be pairwise (ideal-springs, finite-springs) or collective (charges) with different restrictions on the resulting basis networks, as shown in Figure \ref{figure:basis_graphs}. We also make use of the equivalence of symmetries for complementary graphs---the automorphisms of a graph and its complement are identical---meaning we need only consider the combinations of sparse graphs, from which the full set can be constructed by taking the complements of the basis networks (being careful with self-complementary graphs).

\paragraph{Multiplex isomorphism classes} To generate multi-interaction networks we join basis networks together to form a multiplex network. The set of multiplex networks resulting from all the combinations of basis networks, and all the ways of joining them up, can be partitioned into multiplex isomorphism classes. Just as for the basis networks, these can be considered as the meaningfully `unique networks’. An example of multiplex isomorphism classes partitioning the interaction networks, generated by joining basis networks together, is shown in Figure \ref{figure:purple_graphs}. For multiplex networks to be isomorphic, it is necessary that the layers are separately isomorphic, and as such we are guaranteed to include all non-isomorphic multiplex networks when considering only the combinations of basis networks.

\begin{algorithm}[h]
\SetKwInOut{Input}{Input}\SetKwInOut{Output}{Output}
\Input{Graphs $G_1$, $G_2$ of length $k$ with automorphisms $\{A_1^1,A_2^1,...A^1_N\}$, $\{A_2^1,A_2^2,...A^2_M\}$}
\Output{$S$, the set of non-isomorphic multiplex networks with basis graphs  $G_1$, $G_2$}
List all ways of connecting the graphs as permutations of $k$ labels as $l_p$\;
Instantiate empty lists for the checked, $l_c$, and unchecked, $l_u$, members of an equivalence class and an empty set for the output, $S$\;
 \While{$l_p$ is not empty}{
  move a permutation from $l_p$ to $l_u$\;
  \While{$l_u$ is not empty}{
    move a permutation from $l_u$ to $l_c$\;
    \ForEach{automorphism of $G_1$}{
        apply the automorphism directly to the labels of the latest element of $l_c$\;
        \If{the result is not in $l_u$ and not in $l_c$}{
            move the result from $l_p$ and add it to $l_u$    
        }
    }
    \ForEach{automorphism of $G_2$}{
        apply the automorphism to the label positions of the latest element of $l_c$\;
        \If{the result is not in $l_u$ and not in $l_c$}{
            move the result from $l_p$ and add it to $l_u$    
        }
    }
  }
  add an element from $l_c$ to $S$ as the representative of the class and empty $l_c$
 }
\caption{Generating multiplex isomorphism classes with automorphisms}
\label{alg:psuedocode}
\end{algorithm}

\paragraph{Fast multiplex isomorphism generation} To understand the sampling distribution over multiplex isomorphism classes we need to generate them. Naively, this can be achieved by generating all possible networks (binary strings over the number of edges), checking that they are multiplex and satisfy force relations, and then performing pairwise isomorphism tests to build groups. We present a new method that exploits the symmetries of the basis networks and the process of combining them.

The key concept is to write the ways of combining basis graphs as permutations of node labels and then make associations between these using automorphisms. We can write all the ways of combining a pair of graphs with labels $(abc...)$ and $(123...)$, respectively, by keeping the second graph fixed and permuting the nodes in the first. By definition, performing an automorphic transformation on node labels in one layer of the multiplex is undetectable in the other layers, and so the resulting permutation of node labels is isomorphic to the original network. This allows us to construct an equivalence class by applying all basis graph automorphisms, grouping the resulting permutations and further applying the automorphisms to the results to form a closed-group. Notably, any automorphism for the overall network must also be an automorphism for every basis graph, and so we do not overlook any transformations. We provide a visualisation of the method in Figure \ref{fig:algo_visual}, psuedocode in Algorithm \ref{alg:psuedocode} and a proof in Appendix \ref{apdx:proof}.

Our method is applied on pre-generated automorphisms (a task handled by multiple existing libraries \cite{darga2008faster}. To combine a third graph, we can flatten the representative multiplex of $S$ where the automorphisms of the flattened graphs are given by the automorphisms that exist in both basis networks for the given node pairing (permutation). The flattened graph can then be passed as input itself.

\begin{figure*}
\begin{center}
    \includegraphics[trim=51 95 53 13, clip, width=\textwidth]{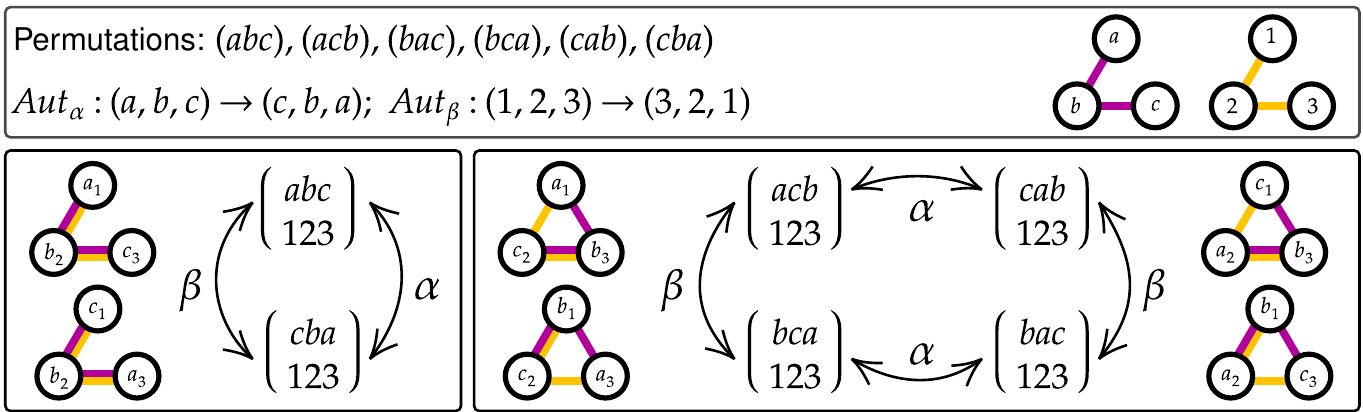}
\end{center}
    
    \caption{Two basis networks are combined to form multiplex networks using our automorphism method. Permutations of $(abc)$ are implicitly connected to the statically ordered $(123)$, which accounts for all unique vertex-aligned ways of connecting the graphs (top). Applying automorphic transformations $\alpha$ and $\beta$ groups permutations into isomorphism equivalence classes (left, right) and leaves non-isomorphic networks separate. Because we keep the $123$-basis network fixed, $\beta$ manifests as exchanging the \textit{first} and \textit{third} elements of the permutation string.}
    \label{fig:algo_visual}
\end{figure*}

\subsection{Sampling biases and data leaks}
Given we are now able to efficiently generate the set of non-isomorphic multiplex networks for a group of interaction types, we turn our attention to the sampling distribution induced by different generation methods and their effects on model performance and evaluation.
\label{section:bias_leak}

\begin{table}[h]
\begin{center}
\begin{small}
\begin{sc}
\begin{adjustbox}{width=\columnwidth}
\begin{tabular}{llccccc}
\toprule
Model & Dataset & MSE20 $/ 10^{-5}$ & Accuracy \\
\midrule
\multirow{2}{*}{fNRI} & Train-ER &19.61$\pm$0.56&0.575$\pm$0.059\\ 
& Train-Uniform &$\mathbf{16.45\pm1.03}$&0.553$\pm$0.019 \\

\midrule
\multirow{2}{*}{NRI} & Train-ER
&428.55$\pm$20.18&0.565$\pm$0.071\\ 
& Train-Uniform
&$\mathbf{376.44\pm8.95}$&0.583$\pm$0.053\\ 
\bottomrule
\end{tabular}
\end{adjustbox}
\end{sc}
\end{small}
\end{center}
\caption{fNRI and NRI performance on the Train-ER and Train-Uniform datasets with ideal-spring, charge, finite-spring interactions. The ER sampling biases affects the predictive performance of the models.}
\label{table:ER_Uniform}
\end{table}

\begin{figure}
\begin{center}
\includegraphics[trim = 5 55 212 5, clip,width=\columnwidth]{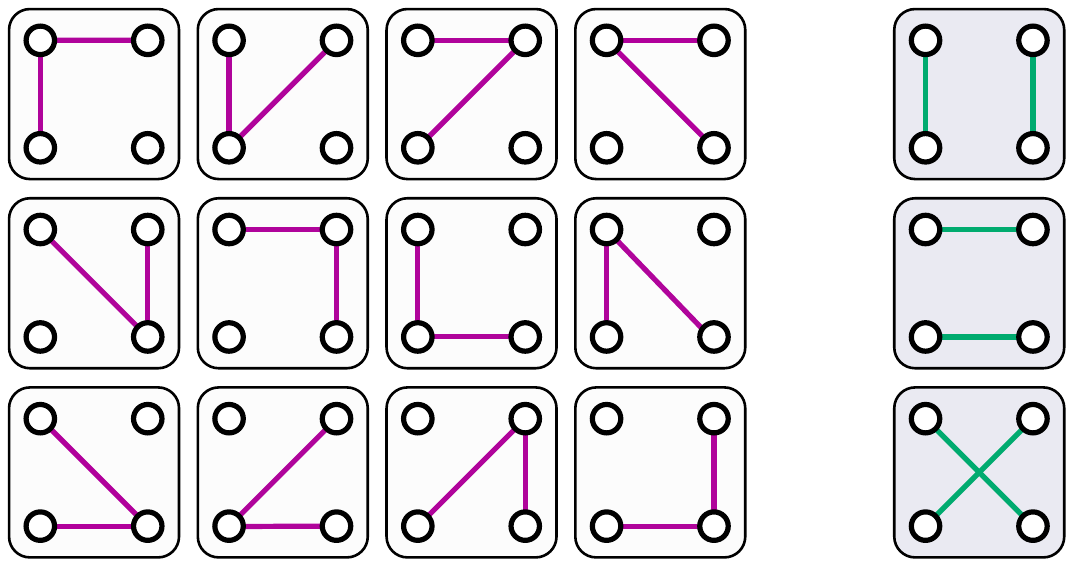}
\end{center}

    \caption{Ways of arranging two edges between four nodes. There are 6 edge positions and 15 (equally-likely) ways of arranging 2 edges on them (6-choose-2). Of these, 12 are in one equivalence class (unshaded, purple) and 3 are in the other (shaded, green). Other factors being equal, the first equivalence class is four times more likely to be sampled than the second in the Original-ER generation procedure.}
    \label{fig:arrangement_bias}
    
\end{figure}

\begin{figure}[ht]
\centering
\includegraphics[width=0.925\columnwidth]{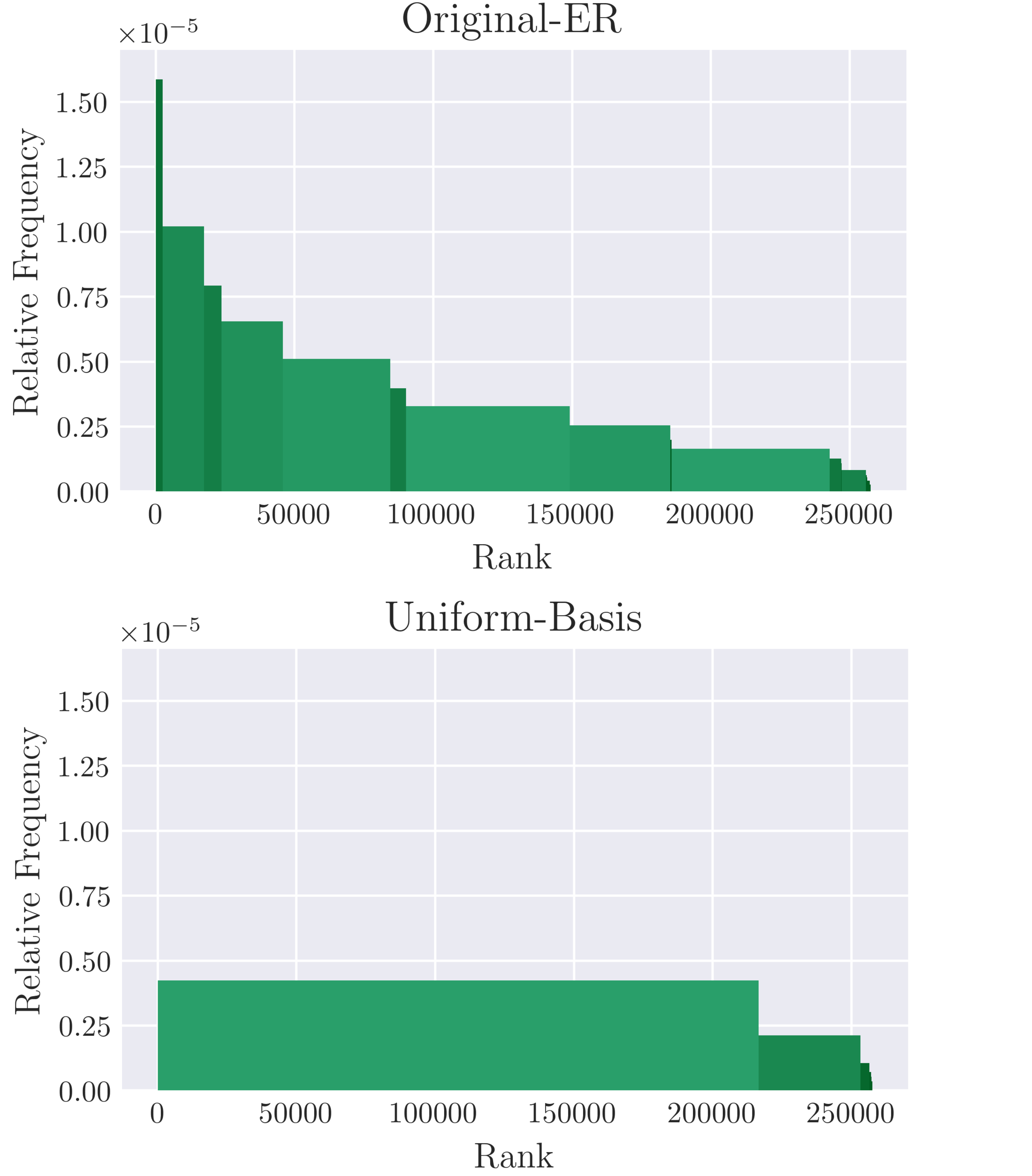}

\caption{Rank-frequency plot of multiplex isomorphism class relative sampling frequencies under the Original-ER generation method (left) and Uniform-Basis method (right). Isomorphism classes with joint-rank are grouped with group colour added to aid visualisation. This shows the strong sampling biases that arise as a consequence of the methods used to generate interaction networks: some equivalence classes are highly prioritised while others are effectively never sampled. The ratio of most-to-least likely is 581:1 for Original-ER and 120:1 for Uniform-Basis.}
\label{figure:sampling_distr}

\end{figure}

\paragraph{Not all networks are created equally} \citet{Kipf:NRI} generate interaction networks with Bernoulli sampling over edges for pairwise interactions and Bernoulli sampling over nodes for collective interactions, a process that is inherited by \citet{Webb:fNRI}. Sampling graphs with a Bernoulli distribution on the presence of edges is commonly known as Erdős–Rényi (ER) sampling and we will refer to this generation procedure as Original-ER. The total number of edges or interacting-nodes follow a binomial distribution and there is a second bias arising for pairwise interactions from their arrangement, as shown in Figure \ref{fig:arrangement_bias}. We also consider a second generation method where basis network isomorphism classes are sampled uniformly, Uniform-Basis, that removes the arrangement bias. By propagating the distribution these methods induce over basis network sampling frequencies, we can produce the relative frequency of the full multiplex network isomorphism classes, shown in Figure \ref{figure:sampling_distr}. We find strong sampling biases exist for both methods, with the most-to-least-likely ratio exceeding 100:1 in each case. 

We compare the performance of models that are trained on training sets generated by ER sampling and uniform sampling of the multiplex isomorphism classes. The latter removes all sampling biases on the multiplex isomorphism classes. The validation and test sets are generated by uniform sampling of the multiplex isomorphism classes, and are identical between both datasets, which will be referred to as Train-ER and Train-Uniform respectively. The results in Table \ref{table:ER_Uniform} show that ER bias reduces the performance on both the models.

\begin{table}[h!]
\begin{center}
\begin{small}
\begin{sc}
\begin{adjustbox}{width=\columnwidth}
\begin{tabular}{llccccc}
\toprule
Model & Dataset & MSE20 $/ 10^{-5}$ & Accuracy \\
\midrule
\multirow{2}{*}{fNRI} & Original-ER      & $\mathbf{21.19\pm 0.53}$ & $\mathbf{0.609\pm0.051}$\\
& Rejection-ER     & $24.82\pm1.83$ & $0.525\pm0.010$ \\
\midrule
\multirow{2}{*}{NRI} & Original-ER      &$\mathbf{24.49 \pm 0.34}$ & $\mathbf{0.604 \pm 0.057}$  \\
&Rejection-ER     &$26.10 \pm 0.48$  &$0.507 \pm 0.059$ \\
\bottomrule
\end{tabular}
\end{adjustbox}
\end{sc}
\end{small}
\end{center}
\caption{fNRI and NRI performance on the Original-ER and Rejection-ER datasets with ideal-spring, charge, finite-spring interactions. Isomorphism leakage in the Original-ER sampling leads to performance overestimation.}
\label{table:ER_ERU}
\end{table}

\begin{table*}[h]

\begin{center}
\begin{small}
\begin{sc}
\begin{adjustbox}{width=\textwidth}
\begin{tabular}{l ccc | ccc | ccc}
\toprule
\multirow{2}{*}{Dataset} & \multicolumn{3}{c}{Initial Conditions} & \multicolumn{3}{c}{Interaction Networks} & \multicolumn{3}{c}{Total Trajectories} \\
 & Train & Val & Test & Train & Val & Test & Train & Val & Test \\
\midrule
Original-ER & Random & Random & Random  & Random & Random & Random  & 50000 & 10000 & 10000 \\
Con-$n$     & $n$ & 22 & 22  & 454 & 454 & 454 & $454 \times n$ & 9988 & 9988  \\ 
Con-111   & 111 & 22 & 22  & 454 & 454 & 454 & 50394 & 9988 & 9988   \\ 
Iso-155   & 155 & 155 & 155 & 324 & 65 & 65 & 50220 & 10075 & 10075 \\ 
Con-Iso   & 155 & 155 & 155 & 324 & 65 & 65 & 50220 & 10075 & 10075 \\ 

\bottomrule
\end{tabular}
\end{adjustbox}
\end{sc}
\end{small}
\end{center}
\caption{Dataset summary for ideal-spring, charge interactions. The number of initial conditions, interaction networks and total number of trajectories for the training, validation and test sets are given.}
\label{table:gen_sum}
\end{table*}

\begin{table}[h!]
\begin{center}
\begin{small}
\begin{sc}
\begin{adjustbox}{width=\columnwidth}
\begin{tabular}{lcccc}
\toprule
Dataset & MSE20 $/ 10^{-5}$ & Accuracy  \\
\midrule
Original-ER  & 10.03 $\pm$ 0.47 & 0.928 $\pm$ 0.008 \\
Con-111   & 14.31 $\pm$ 0.71 & 0.943 $\pm$ 0.005\\
Iso-155   & $8.07 \pm 0.56$ & $0.965 \pm 0.001$ \\
Con-Iso   &9.65$\pm$0.33&$0.534\pm0.003$\\

\bottomrule
\end{tabular}
\end{adjustbox}
\end{sc}
\end{small}
\end{center}
\caption{fNRI performance on the ideal-spring, charge datasets. The fNRI demonstrates generalisation to different initial conditions, multiplex isomorphism classes and both.}
\label{table:gen_result}
\end{table}

\paragraph{Isomorphism leakage} Conventionally, training, validation and test sets are disjoint. Naive generation of the interaction networks however will almost certainly result in some multiplex isomorphism classes being present in the different sets, i.e. leaking to the test set. For two datasets $X_{1}$ and $X_{2}$ with data $x(G)$ generated by some latent graph $G$, we say that there is isomorphism leakage between $X_{1}$ and $X_{2}$ if there exists $x_{1}(G_{1})\in X_{1}$ and $x_{2}(G_{2})\in X_{2}$ where $G_{1}$ and $G_{2}$ are isomorphic. Neither \citet{Kipf:NRI} nor \citet{Webb:fNRI} claim to control for this possibility, and it is not the case that \textit{identical} examples are present across their splits as initial conditions vary, however, by controlling for this facet of variability in isolation we find significant changes in judged model performance. We adopt the exact training scheme used by \citet{Webb:fNRI}\footnote{Both the original NRI and fNRI have made their codebases publicly available, greatly enabling this work.} to compare models trained on datasets produced with the Original-ER method and an adaptation that controls for isomorphism leakage by rejecting test samples from multiplex isomorphism classes present in the training set (Rejection-ER). The results presented in Table \ref{table:ER_ERU} show that the leaky test set judges models to produce better trajectories and more accurately infer interaction relations, thus overestimating performance.

\section{Model testing}\label{section:model}

In light of the previous results, there is a need for a standardised and reproducible isomorphism-aware benchmarking framework to evaluate model performances \cite{Dwivedi:benchmark}. In this section we propose multiple benchmarks and utilise them, to analyse the fNRI. We test for generalisability, focusing on the evaluation of flexibility and robustness over `skill' \cite{Chollet:measure}. We also investigate how the training set distribution affects performance, including varying the sampling frequency of isomorphism classes in the training set.

\paragraph{Measuring generalisation} Considering the uniqueness of interaction networks, we can associate testing on isomorphism classes seen during training with the transductive setup \cite{Yang2016RevisitingEmbeddings}, where the same graph is used in both contexts, and testing on unseen classes with the inductive setup. We can further associate two kinds of generalisation with these cases: to different initial conditions (Con) in the transductive case, and to different interaction networks (Iso) in the inductive case.

Here we focus on the ideal-spring and charge interactions for five particles as the number of multiplex isomorphism classes is small (454). The original work \cite{Kipf:NRI} used what will be referred to as the Original-ER dataset for ideal-spring, charge interactions. This has the same structure as in Section \ref{section:bias_leak}, which also included finite-springs. To test Con and Iso generalisation, and also compare our results with the Original-ER dataset, we propose the Con-$111$ dataset and Iso-$155$ dataset respectively. We also investigate both types of generalisation together using the Con-Iso dataset. The Con-111 dataset contains [454, 454, 454] multiplex isomorphism classes (all of them), with [111, 22, 22] initial conditions (the same set for each multiplex isomorphism class). The Iso-155 dataset partitions the multiplex isomorphism classes between the training, validation and test set such that they do not overlap, each with the same set of 155 initial conditions. The Con-Iso dataset also has the same structure, but all the initial conditions are different. In these datasets the number of initial conditions is chosen such that the total trajectories closely matches that of the Original-ER dataset e.g. $111 \times 454 \approx 50000$. A summary of these datasets and the results are shown in Table \ref{table:gen_sum} and \ref{table:gen_result} respectively.

The fNRI performs well on the Con-111 and Iso-155 datasets and demonstrates generalisation to different initial conditions, multiplex isomorphism classes separately. Perhaps counter-intuitively, the Iso-155 dataset actually outperforms the Con-111 dataset. A plausible explanation is that repeatedly observing the same interaction networks applied to different initial conditions provides a stronger learning signal, which in turn enables superior performance and generalisation. The Con-Iso dataset demonstrates good performance for the mean-squared error, but has a much lower encoding accuracy. This may be due to the model learning an implicit encoding of the trajectories that isn't interaction network based, as opposed to the explicit one which the encoding accuracy measures.

\begin{figure}[h!]
\begin{center}
\centerline{\includegraphics[width=0.975\columnwidth]{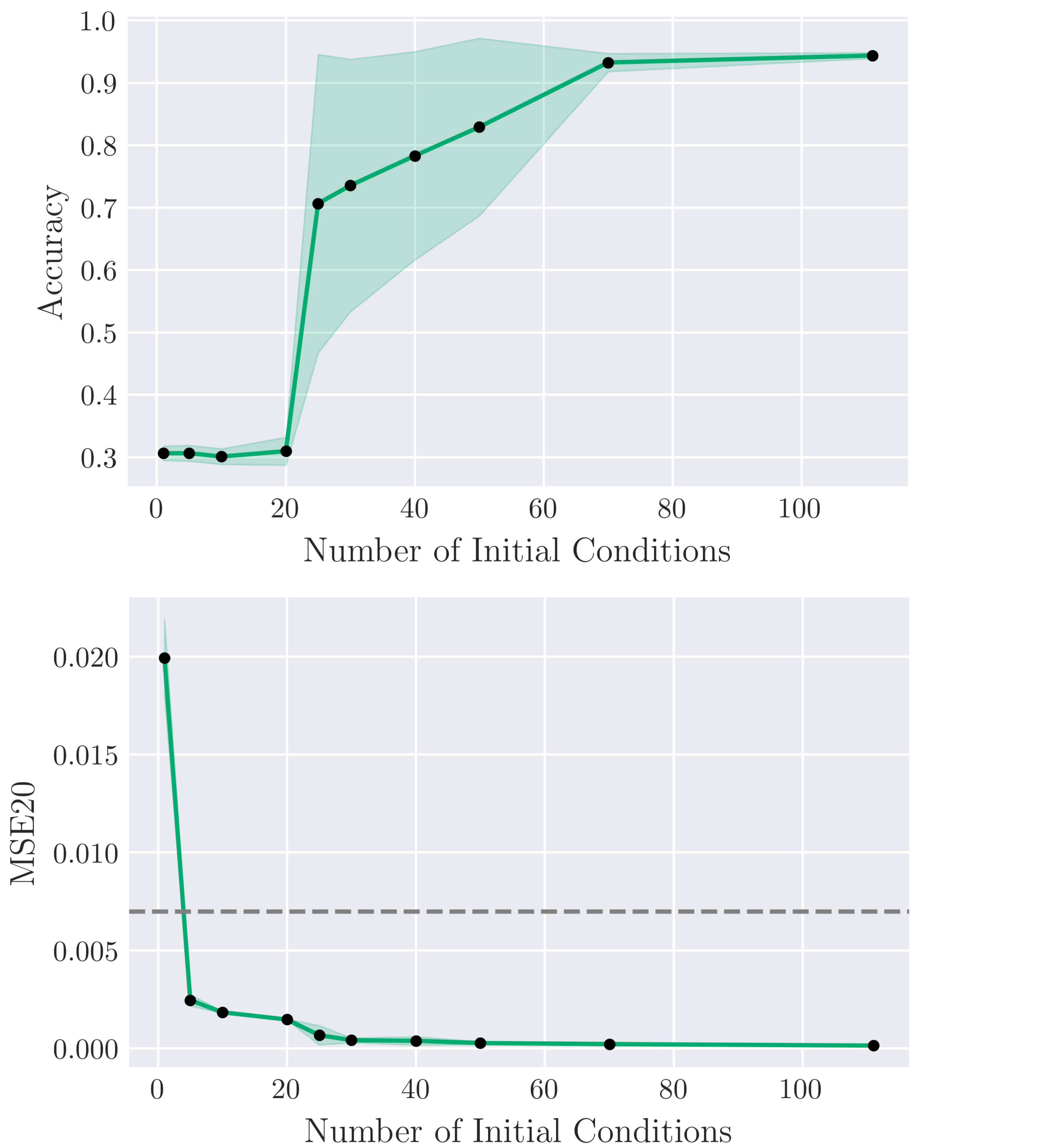}}
\caption{Performance of the fNRI vs. the number of initial conditions for each isomorphism class in the training set for ideal-spring, charge interactions. The encoding accuracy (left), higher is better. This shows a sharp rise at around 20 initial conditions. The mean-squared error (MSE) for 20 time-steps (right), lower is better. The dashed grey line shows the MSE for stationary particles. The MSE shows a significant drop at around 20 initial conditions.}
\label{figure:EoF}
\end{center}
\end{figure}

\begin{figure*}[h!]
\begin{center}
\centerline{\includegraphics[clip, width=0.8\textwidth]{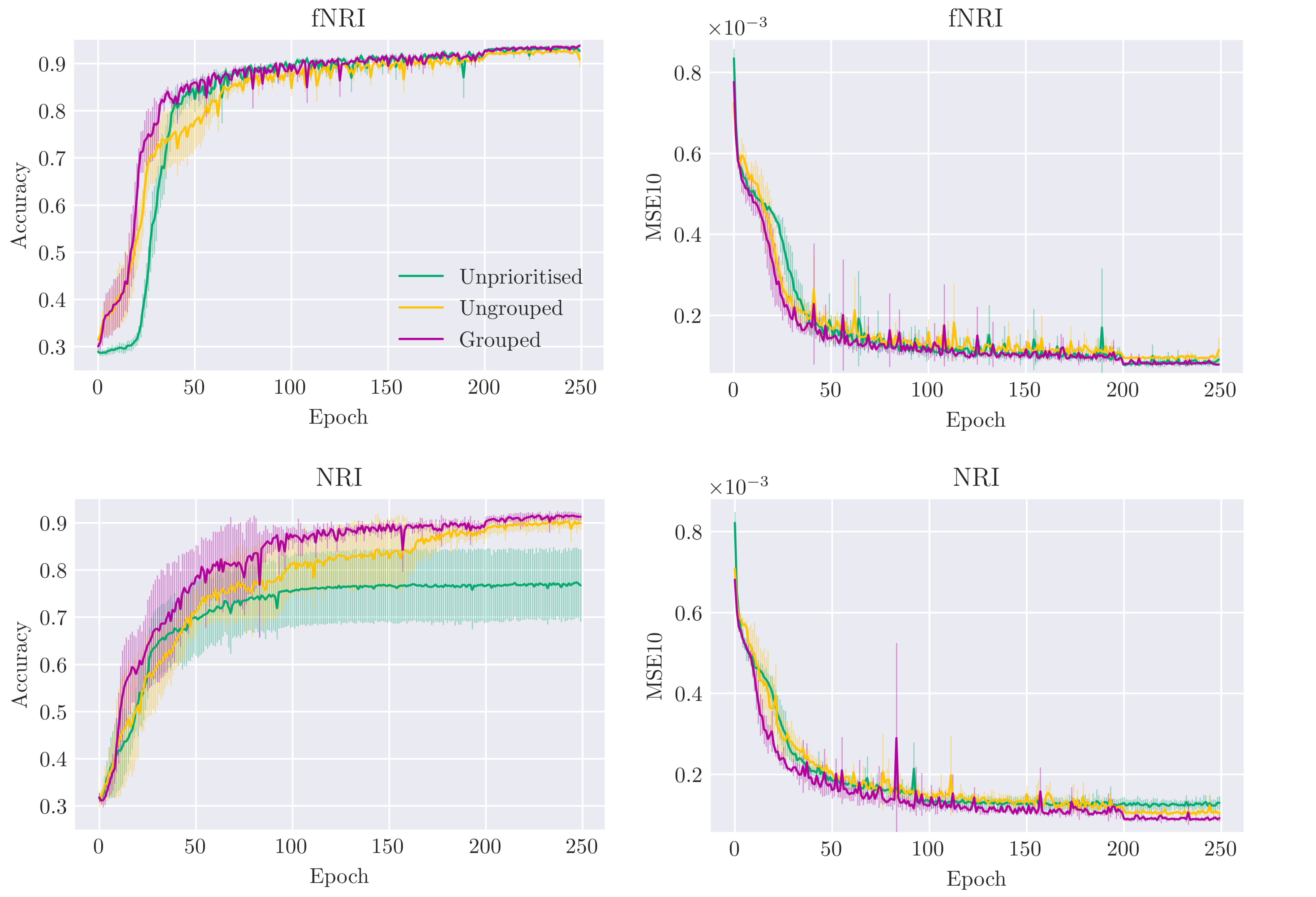}}
\caption{Mean validation performance for unprioritised and prioritised sampling with and without grouping by multiplex isomorphism classes of ideal-spring, charge interactions. Prioritised sampling with grouping reliably increases the learning rate and the performance of both the fNRI and NRI.}
\label{figure:priority}
\end{center}
\end{figure*}

\paragraph{Few-Shot learning} Though state-of-the-art performance in many machine learning tasks is achieved with large amounts of labelled data, there are many domains in which it is impractical or overly costly to gather sufficient data. In such a setting, few-shot learning algorithms can be employed to make best use of what is available  \cite{Wang:FSL}. The effect of the frequency of occurrence of each multiplex isomorphism class in the training set is explored in this section to investigate the fNRI's capacity for few-shot learning. The Con-$n$ datasets are used, which have the same structure as the Con-111 dataset, where each multiplex isomorphism class is seen $n$-times per epoch. Figure \ref{figure:EoF} shows how model performance improves as the number of initial conditions, $n$, is increased.

% \begin{figure}[h!]
% \begin{center}
% \centerline{\includegraphics[width=\columnwidth]{figures/acc_mse20.pdf}}
% \caption{Performance of the fNRI vs. the number of initial conditions for each isomorphism class in the training set for ideal-spring, charge interactions. The encoding accuracy (left), higher is better. This shows a sharp rise at around 20 initial conditions. The mean-squared error (MSE) for 20 time-steps (right), lower is better. The dashed grey line shows the MSE for stationary particles. The MSE shows a significant drop at around 20 initial conditions.}
% \label{figure:EoF}
% \end{center}
% \end{figure}

As expected, the performance of the fNRI increases and then plateaus as the number of initial conditions increase. There is a rise in performance at around 20 initial conditions, which is associated with the increase in encoding accuracy. This shows that there is a threshold sampling frequency of isomorphism classes for successful learning. Curiously, there is a large drop in the mean-squared error below the threshold frequency. Again, we believe that the model may be learning to encode an implicit representation of the interaction network, which allows for the prediction of the few trajectories that are present in the smaller datasets with lower number of initial conditions. Once we pass the threshold frequency, it may be that the explicit representation is required to capture the entire dataset, hence the increase in encoding accuracy and consequently the decrease of the mean-squared error.

\section{Prioritised sampling}\label{subsec:prioritised}
In this section, we use a simple prioritisation scheme to analyse the benefits isomorphism-awareness can have on training speed and final performance. The likelihood of selecting an example for training is proportional to the exponentially weighted average of the historic model error on that sample. The historic error can also be grouped by multiplex isomorphism class. The performance of unprioritised and prioritised sampling with and without grouping is shown in Figure \ref{figure:priority}.

% \begin{figure*}[h!]
% \begin{center}
% \centerline{\includegraphics[clip, width=0.1\textwidth]{figures/priority.pdf}}
% \caption{Mean validation performance for unprioritised and prioritised sampling with and without grouping by multiplex isomorphism classes of ideal-spring, charge interactions. Prioritised sampling with grouping reliably increases the learning rate and the performance of both the fNRI and NRI.}
% \label{figure:priority}
% \end{center}
% \end{figure*}

Prioritised sampling in the fNRI reliably shifts the learning curve to lower epochs, increasing the learning rate. Without grouping by multiplex isomorphism classes, the fNRI converges to lower performances. Prioritised sampling also improves performance in the NRI, again to a lesser extent without grouping. This demonstrates that isomorphism-awareness can be used to improve model performances.

\section{Conclusions}
We have analysed multiplex isomorphism classes in the context of learning to model multi-interaction systems. On the basis of our analysis, we have shown that the performance of models on this task has been overestimated, particularly with regards to generalisation. To remedy this problem we have proposed and evaluated new benchmarking datasets. Through experiments with these new benchmarks we show under what conditions neural relational inference models can be expected to learn and generalise well. We also present results on prioritised sampling in a training procedure that parallels the scientfic process. Finally, we have presented, and proven, an efficient new method for generating multiplex isomorphism classes for this context that makes further work in this area practical and accessible.

\clearpage

\section*{Ethics Statement}
Our work is concerned primarily with foundational results in graph theory and their implications for training and evaluation for similarly foundational problems in systems of interacting particles. For this reason we consider there to be few foreseeable broader impacts, though we address the potential application of these models to human networks.

The NRI \cite{Kipf:NRI} presents results applying the model to the motion of basketball players and it seems reasonable to consider whether this could be extended to generic motion of people. Firstly, the application is more narrow than it first appears, predicting motion during an artificially constrained phase of play (a pick-and-roll) and the model is only weakly able to \textit{reconstruct} player trajectories even in this scenario. Secondly, there is a scaling issue with the current system that requires $O(N^2)$ relations to be considered which has not been resolved (limiting application to larger groups). Thirdly, it is unclear how the data collection to enable this application could be performed without also having the infrastructure to make it redundant---if you have high quality segmented overhead video footage of citizens, why do you need a model to tell you how they will move?

A second consideration may be that the model could be adopted to track how individuals interact and `move' online. Whilst it would be interesting to investigate whether these models can be used in a discrete, non-Euclidean space, current work is limited to particles moving in a continuous, low-dimensional, Euclidean space only, and it is far from obvious how to solve key challenges to adapting to this new task.

\bibliography{bib}

\clearpage

\appendix

\section{Proof for the multiplex graph isomorphism algorithm}
\label{apdx:proof}
To show the algorithm works, we need to show given one representative of an isomorphism class, it generates all isomorphic layered graphs.

Suppose we have graphs $G_0,...,G_r$ as our basis graphs, together with embeddings $f_i : V(G_i)\rightarrow V$ a common vertex set. We can identify this vertex set $V$ with the vertex set $V(G_0)$, and choose to do so. We can therefore assume we have bijections $h_i : V(G_i)\rightarrow V(G_0)$ by post-composing with $f_0^{-1}$. 

Suppose that $g_0,...,g_r$ give an isomorphic embedding to the $f_i$. Let $\gamma$ be an isomorphism which witnesses the layered graphs are isomorphic, i.e. there is an edge between $\gamma g_i (v)$ and $\gamma g_i (w)$ iff there is an edge between $f_i(v)$ and $f_i(w)$. Since all of these maps are bijections, we deduce $f_i^{-1}\gamma g_i$ exists and is an automorphism of $G_i$ for each $i$. Call this map $\tau_i$. 

Note that $\gamma^{-1}f_i\tau_i = g_i \forall i$. 
Returning to the $h$ maps, we want to transform the $f_0^{-1}f_i$ into the $g_0^{-1}g_i$ by postcomposition of a common isomorphism of $G_0$, and precomposition by isomorphisms of $G_i$.  Using our previous map, we observe $g_0^{-1}g_i = \tau_0^{-1}g_0^{-1}\gamma\gamma^{-1}f_i \tau_i = \tau_0^{-1}f_0^{-1}f_i\tau_i$. This has the same form as in our algorithm, and hence we must obtain every isomorphic embedding.

\section{Other investigations}
In this section we present other investigations that were conducted on the fNRI using isomophism-aware benchmarks. This includes identifying which interaction types are most important for training, the effect of not including all isomorphism classes in the training set, and measuring generalisation for three interactions (as opposed to two interactions in Section \ref{section:model}).

\subsection{Training essentials}
\label{section:essentials}
The importance of each interaction type in the training set for the fNRI on the ideal-spring, charge dataset is investigated here. The datasets used are: 

\begin{itemize}
    \item Extrapolate Charges to High (XCH)
    \item Extrapolate Charges to Low (XCL)  
    \item Interpolate Charges (IC)
    \item Extrapolate Springs to High (XSH)
    \item Extrapolate Springs to Low (XSL)
    \item Interpolate Springs (IS)
\end{itemize}

Using the XCH dataset as an example, the interaction networks are split into high charge and low charge groups. The training set is comprised of the low charges. The validation and test set is comprised of the high charges. Each has [50, 22, 22] initial conditions. The same logic and number of initial conditions apply to the other datasets. 

\begin{table*}[h]
\begin{center}
\begin{small}
\begin{sc}
\begin{tabular}{lcccr}
\toprule
Dataset & MSE20 $/ 10^{-5}$ & Accuracy & I-Spring & Charge \\
\midrule
XSH     & 96.48$\pm$ 64.26& 0.567$\pm$ 0.288& 0.688$\pm$ 0.181& 0.779$\pm$ 0.250 \\
XSL     & 105.40 $\pm$ 38.21 & 0.568 $\pm$ 0.278& 0.724$\pm$ 0.187& 0.749$\pm$ 0.191\\
IS      & 99.33$\pm$ 9.29& 0.380$\pm$0.118& 0.636 $\pm$0.082& 0.563$\pm$0.068\\
\midrule
XCH     &$230.45\pm9.56$&$0.384\pm0.063$&$0.626\pm0.074$&$0.572\pm0.025$ \\
XCL     &$\mathbf{32.05\pm17.62}$&$0.605\pm0.242$&$0.708\pm0.201$&$\mathbf{0.836\pm0.095}$ \\ 
IC      &$73.49\pm27.80$&$0.412\pm0.144$&$0.629\pm0.171$&$0.649\pm0.170$\\ 
\bottomrule
\end{tabular}
\end{sc}
\end{small}
\end{center}
\caption{fNRI performance on extrapolation/interpolation datasets with ideal-spring (top) and charge (bottom) interactions. The fNRI has comparable performance on all the ideal-spring datasets and performs the best on the XCL dataset for the charges.}
\label{table:spring}

\end{table*}

\begin{figure}[h!]
\begin{center}
\centerline{\includegraphics[width=\columnwidth]{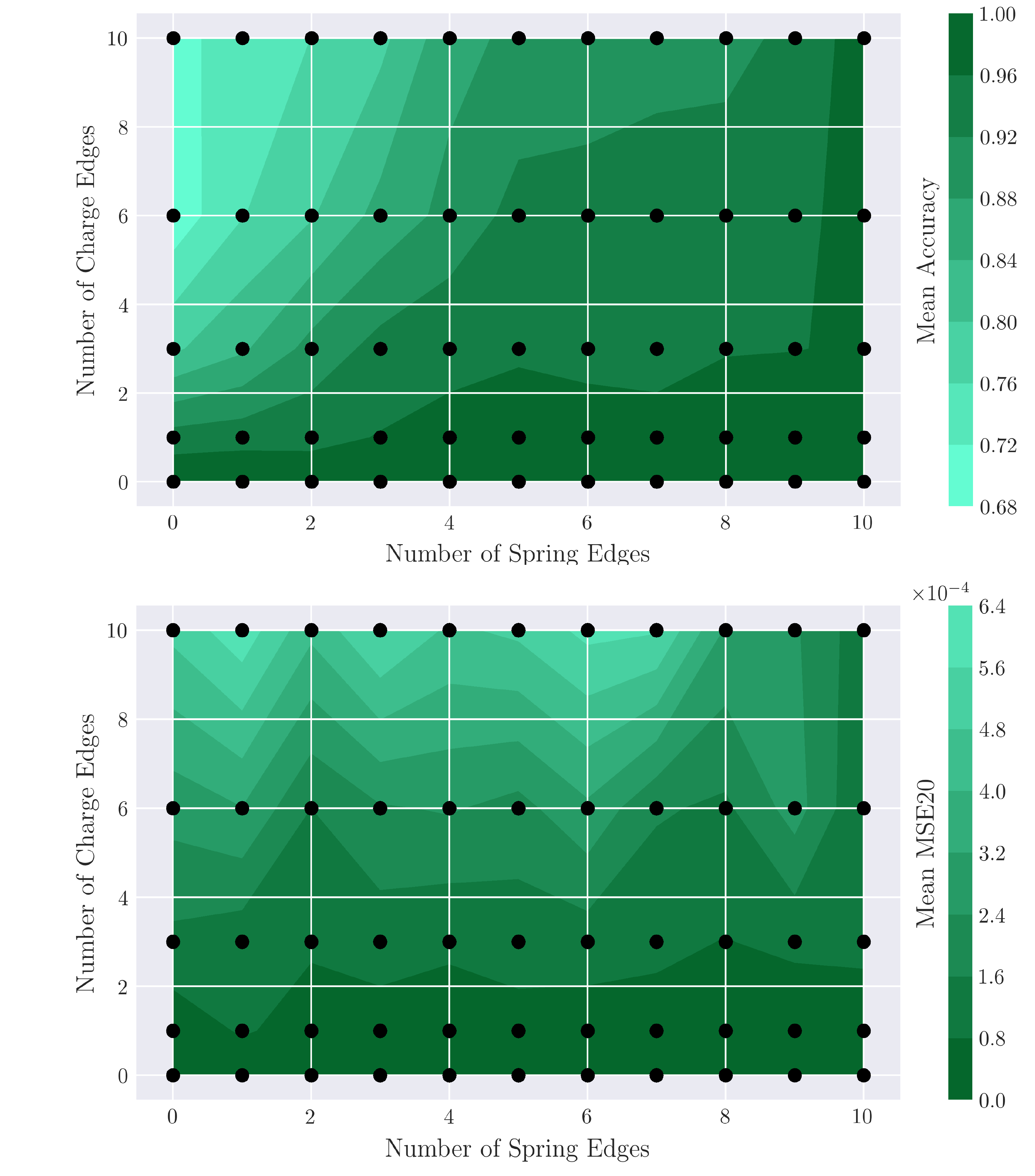}}
\caption{Encoding accuracy (left) and mean-squared error (right) for 20 time-steps of the predicted trajectories vs. different combinations of interaction types in the interaction network. The colour-scale is chosen such that light colours are associated with worse performance. The high charge interactions are associated with lower performance for both the accuracy and the mean-squared error. The performance is roughly constant for the mean-squared error, with respect to variations in the number of springs, whereas it decreases with decreasing number of springs for the accuracy.}
\label{figure:competence}
\end{center}
\end{figure}

The results in Table \ref{table:spring} show that the fNRI has comparable performance for the spring datasets, and performs the best on the XCL dataset for the charges. Training on high charges seems to allow for better generalisation to lower charges whereas there seems to be no preference for the springs. To gain insight into these results, we identify the `difficulty' of each interaction type. To do this we trained the fNRI on the Con-111 dataset and partitioned the test set by interaction type. The results are shown in Figure \ref{figure:competence}.

According to the reconstruction error (MSE20), the most difficult interactions are the high charge interaction networks, which seems to become easier for no springs and high springs.  This is expected for the no spring case and the high spring (purely attractive) case may be explained by the clumping of particles which may cause the fNRI to `cheat' and predict the centre of mass motion of the particles. Besides this, the spring difficulty seems to be roughly constant for each number of charge-edges. The difficulty, according to the encoding accuracy, is highest for high charges and low springs. This may explain the results on the extrapolation/interpolation datasets. According to the reconstruction error, the training set `difficulty' should be around the same for the spring datasets, whereas the XCL dataset should have the most difficult training set. This may suggest that training on interactions the model find the most difficult may generalise better to easier interactions, and not the other way around. Note that the fNRI only has access to the reconstruction error (in the loss function) and not the encoding accuracy.

\subsection{The effect of sub-sampling}
\label{section:ER_revisit}
%In this section we remove the effect of ER bias on the fNRI performance on the ER dataset with SCfS interactions. The SCfS interaction networks have around 250000 multiplex isomorphism classes. The SCfS datasets generated by ER sampling in Section \ref{section:isomorphism_leakage} will therefore not cover all the multiplex isomorphism classes, cf. the SC interactions where ER sampling does cover all of them. Let us first examine the effect of this sub-sampling of the multiplex isomorphism classes in the training set for the SC interactions.  Here we compare the performance of Con-$n$ datasets, which contain all the multiplex isomorphism classes, with the performance on the Sub-Con-$n$ dataset. This dataset removes some multiplex isomorphism classes from the training set of the Con-$n$ dataset such that there are [324, 454, 454] multiplex isomorphism classes. The validation and test sets are identical between these two datasets. 

In this section we consider the effect of not including all the possible multiplex isomorphism classes in the training set, for ideal-spring charge interactions. We compare the performance of Con-$n$ datasets, which contain all the multiplex isomorphism classes, with the performance on the Sub-Con-$n$ dataset. This dataset removes some multiplex isomorphism classes from the training set of the Con-$n$ dataset such that there are [324, 454, 454] multiplex isomorphism classes. The validation and test sets are identical between these two datasets.

\begin{figure}[h!]
\begin{center}
\centerline{\includegraphics[width=\columnwidth]{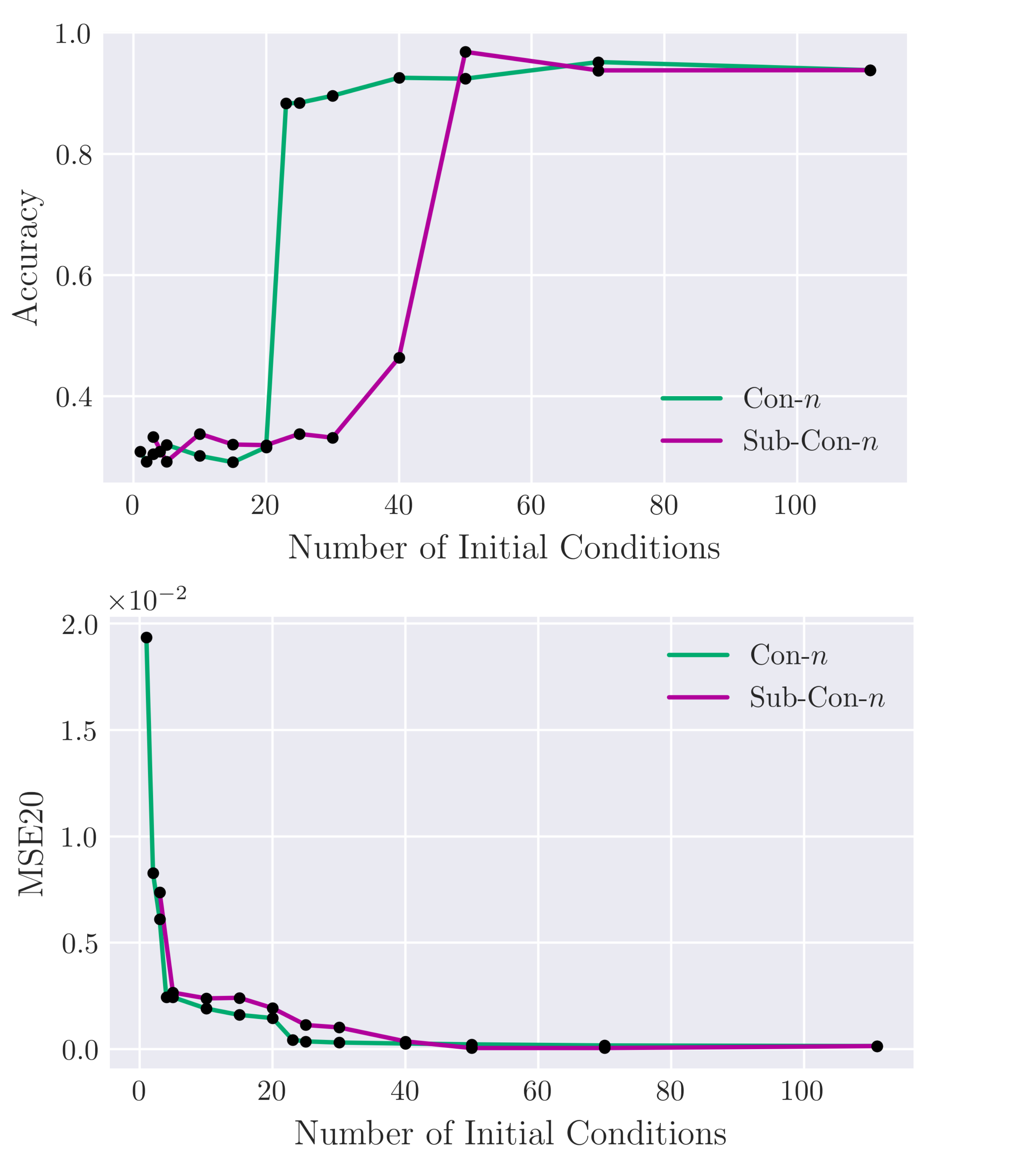}}
\caption{A comparison between the fNRI performance on the Con-$n$ and Sub-Con-$n$ datasets. (left) The encoding accuracy (higher is better). (right) The predicted trajectory mean-squared error for 20 time steps (lower is better). The Sub-Con-$n$ dataset curve approximately has the same behaviour as for the Con-$n$ dataset curve, but shifted to higher number of initial conditions.}
\label{figure:EoF_sub}
\end{center}
\end{figure}

The Sub-Con-$n$ datasets generally show the same behaviour as the Con-$n$ datasets. It performs worse, but eventually catches up to the performance on the Con-$n$ dataset. The threshold frequency of learning has been shifted from around 20 initial conditions to around 40 initial conditions. This suggests that sub-sampling increases the threshold frequency of learning.

\subsection{Measuring generalisation for three interactions}

Here we focus on the ideal-spring, charge and finite-spring interactions for five particles, as opposed to the case for ideal-spring and charge interactions in Section \ref{section:model}. We use the Con-$111$, Iso-$155$ and Con-Iso dataset as before, but for ideal-spring, charge, finite-spring interactions. In this case, the number of multiplex isomorphism classes is large (over 250,000) and we are no longer constrained to using just 454 of them, as in the case for ideal-spring, charge interactions. We keep the same structure for the Con-$111$ and Iso-$155$ datasets, but for the con-iso dataset we use [454, 454, 454] interaction networks, all from different multiplex isomorphism classes. A summary of the results are shown in Table \ref{table:SCF_gen}.

% \begin{table*}[h!]
% \begin{center}
% \begin{small}
% \begin{sc}
% \begin{adjustbox}{width=\textwidth}
% \begin{tabular}{lccccc}
% \toprule
% Dataset & MSE20 $/ 10^{-5}$ & Accuracy & I-Spring Accuracy & Charge Accuracy & F-Spring Accuracy \\
% \midrule
% Original-ER &$21.19\pm 0.53$ & $0.609\pm0.051$ &  $0.866 \pm 0.025$ & $0.975\pm0.010$ & $0.673\pm0.044$ \\
% Con-111&  $37.59\pm1.07$& $0.548\pm0.013$ & $0.831\pm0.015$ & $0.978\pm0.003$ & $0.619\pm0.006$ \\
% Iso-155 & $15.68\pm0.34$ & $0.604\pm0.024$ & $0.873\pm0.019$ & $0.979\pm0.004$ & $0.660\pm0.017$ \\
% Con-Iso & $31.84\pm0.16$ & $0.567\pm0.019$ & $0.849\pm0.016$ & $0.984\pm0.001$ & $0.630\pm0.012$ \\

% \bottomrule
% \end{tabular}
% \end{adjustbox}
% \end{sc}

% \end{small}
% \end{center}
% \caption{fNRI performance on the ideal-spring, charge, finite-spring datasets. The fNRI has the best performance on the Iso-155 dataset, and the worst on the Con-111 dataset. Recall that the Original-ER dataset overestimates the performance due to isomorphism leakage.}
% \label{table:SCF_gen}
% \end{table*}

The results are qualitatively similar to the results for the ideal-spring, charge interactions in Section \ref{section:model}, with the fNRI performing the best on the Iso-155 dataset and the worst on the Con-111 dataset. This shows that the fNRI generalises better to different graphs, compared to different initial conditions. Again, a plausible explanation is that repeatedly observing the same interaction networks applied to different initial conditions provides a strong learning signal.

\begin{table*}[h!]
\begin{center}
\begin{small}
\begin{sc}
\begin{adjustbox}{width=\textwidth}
\begin{tabular}{lccccc}
\toprule
Dataset & MSE20 $/ 10^{-5}$ & Accuracy & I-Spring Accuracy & Charge Accuracy & F-Spring Accuracy \\
\midrule
Original-ER &$21.19\pm 0.53$ & $0.609\pm0.051$ &  $0.866 \pm 0.025$ & $0.975\pm0.010$ & $0.673\pm0.044$ \\
Con-111&  $37.59\pm1.07$& $0.548\pm0.013$ & $0.831\pm0.015$ & $0.978\pm0.003$ & $0.619\pm0.006$ \\
Iso-155 & $15.68\pm0.34$ & $0.604\pm0.024$ & $0.873\pm0.019$ & $0.979\pm0.004$ & $0.660\pm0.017$ \\
Con-Iso & $31.84\pm0.16$ & $0.567\pm0.019$ & $0.849\pm0.016$ & $0.984\pm0.001$ & $0.630\pm0.012$ \\

\bottomrule
\end{tabular}
\end{adjustbox}
\end{sc}

\end{small}
\end{center}
\caption{fNRI performance on the ideal-spring, charge, finite-spring datasets. The fNRI has the best performance on the Iso-155 dataset, and the worst on the Con-111 dataset. Recall that the Original-ER dataset overestimates the performance due to isomorphism leakage.}
\label{table:SCF_gen}
\end{table*}

\begin{table*}[h]
\begin{center}
\begin{small}
\begin{sc}
\begin{adjustbox}{width=\textwidth}
\begin{tabular}{llccccc}
\toprule
Model & Dataset & MSE20 $/ 10^{-5}$ & Accuracy & I-Spring & Charge & F-Spring \\
\midrule
\multirow{2}{*}{fNRI} & Train-ER &19.61$\pm$0.56&0.575$\pm$0.059&0.852$\pm$0.031&0.978$\pm$0.004&0.637$\pm$0.052\\ 
& Train-Uniform &$\mathbf{16.45\pm1.03}$&0.553$\pm$0.019 &0.860$\pm$0.012 &0.981$\pm$0.004 &0.611$\pm$0.021 \\

\midrule
\multirow{2}{*}{NRI} & Train-ER
&428.55$\pm$20.18&0.565$\pm$0.071&0.865$\pm$0.018&0.937$\pm$0.072&0.644$\pm$0.038\\ 
& Train-Uniform
&$\mathbf{376.44\pm8.95}$&0.583$\pm$0.053&0.890$\pm$0.006&0.929$\pm$0.063&0.665$\pm$0.018\\ 
\bottomrule
\end{tabular}
\end{adjustbox}
\end{sc}
\end{small}
\end{center}
\caption{fNRI and NRI performance on the Train-ER and Train-Uniform datasets with ideal-spring, charge, finite-spring interactions. The ER sampling biases affects the predictive performance of the models.}

\end{table*}

\begin{table*}[h]
\begin{center}
\begin{small}
\begin{sc}
\begin{adjustbox}{width=\textwidth}
\begin{tabular}{llccccc}
\toprule
Model & Dataset & MSE20 $/ 10^{-5}$ & Accuracy & I-Spring & Charge & F-Spring \\
\midrule
\multirow{2}{*}{fNRI} & Original-ER      & $\mathbf{21.19\pm 0.53}$ & $\mathbf{0.609\pm0.051}$ &  $\mathbf{0.866 \pm 0.025}$ & $0.975\pm0.010$ & $\mathbf{0.673\pm0.044}$ \\
& Rejection-ER     & $24.82\pm1.83$ & $0.525\pm0.010$ & $0.814\pm 0.010$ & $0.973\pm0.003$ & $0.599\pm0.005$ \\
\midrule
\multirow{2}{*}{NRI} & Original-ER      &$\mathbf{24.49 \pm 0.34}$ & $\mathbf{0.604 \pm 0.057}$  &$\mathbf{0.885 \pm 0.001}$  &$\mathbf{0.922 \pm 0.090}$  & $\mathbf{0.687 \pm 0.009}$  \\
&Rejection-ER     &$26.10 \pm 0.48$  &$0.507 \pm 0.059$  &$0.869 \pm 0.004$  &$0.793 \pm 0.062$  & $0.659 \pm 0.021$  \\
\bottomrule
\end{tabular}
\end{adjustbox}
\end{sc}
\end{small}
\end{center}
\caption{fNRI and NRI performance on the Original-ER and Rejection-ER datasets with ideal-spring, charge, finite-spring interactions. Isomorphism leakage in the Original-ER sampling leads to performance overestimation.}

\end{table*}

\begin{table*}[h!]
\begin{center}
\begin{small}
\begin{sc}
\begin{adjustbox}{width=\textwidth}
\begin{tabular}{lcccc}
\toprule
Dataset & MSE20 $/ 10^{-5}$ & Accuracy & Spring Accuracy & Charge Accuracy \\
\midrule
Original-ER  & 10.03 $\pm$ 0.47 & 0.928 $\pm$ 0.008 & $0.980 \pm 0.001$ & 0.959 $\pm$ 0.019 \\
Con-111   & 14.31 $\pm$ 0.71 & 0.943 $\pm$ 0.005 & 0.971 $\pm$ 0.002 & 0.970 $\pm$ 0.004 \\
Iso-155   & $8.07 \pm 0.56$ & $0.965 \pm 0.001$ & $0.983 \pm 0.002$ & $0.981 \pm 0.002$ \\
Con-Iso   &9.65$\pm$0.33&$0.534\pm0.003$&$0.706\pm0.004$&$0.682\pm0.005$\\

\bottomrule
\end{tabular}
\end{adjustbox}
\end{sc}
\end{small}
\end{center}
\caption{fNRI performance on the ideal-spring, charge datasets. The fNRI demonstrates generalisation to different initial conditions, multiplex isomorphism classes and both.}

\end{table*}

\end{document}